\documentclass[10pt,twocolumn,letterpaper]{article}

\usepackage[pagenumbers]{cvpr} 
\usepackage{graphicx}
\usepackage{amsmath}
\usepackage{amssymb}
\usepackage{booktabs}
\usepackage[table]{xcolor}
\usepackage{appendix}

\usepackage[pagebackref,breaklinks,colorlinks]{hyperref}

\usepackage[capitalize]{cleveref}
\crefname{section}{Sec.}{Secs.}
\Crefname{section}{Section}{Sections}
\Crefname{table}{Table}{Tables}
\crefname{table}{Tab.}{Tabs.}

\usepackage{booktabs}
\newcommand{\squeezeup}{\vspace{-4mm}}
\newcommand{\squeezeupsmall}{\vspace{-2mm}}

\def\cvprPaperID{16} 
\def\confName{CVPR}
\def\confYear{2023}

\begin{document}

\title{RoSteALS: Robust Steganography using Autoencoder Latent Space}

\author{Tu Bui$^{1}$ \\
\and Shruti Agarwal$^{2}$ \\
\and Ning Yu$^{3}$ \\
\and John Collomosse$^{1,2}$\\
\and \vspace{-1.1cm}\\
$^{1}$University of Surrey, $^{2}$Adobe Research, $^{3}$Salesforce Research\\
{\tt\small t.v.bui@surrey.ac.uk, \{shragarw,collomos\}@adobe.com, ning.yu@salesforce.com}
}
\maketitle


\begin{abstract}
   Data hiding such as steganography and invisible watermarking has important applications in copyright protection, privacy-preserved communication and content provenance. Existing works often fall short in either preserving image quality, or robustness against perturbations or are too complex to train. We propose RoSteALS, a practical steganography technique leveraging frozen pretrained autoencoders to free the payload embedding from learning the distribution of cover images. RoSteALS has a light-weight secret encoder of just 300k parameters, is easy to train, has perfect secret recovery performance and comparable image quality on three benchmarks. Additionally, RoSteALS can be adapted for novel cover-less steganography applications in which the cover image can be sampled from noise or conditioned on text prompts via a denoising diffusion process. Our model and code are available at \url{https://github.com/TuBui/RoSteALS}.
\end{abstract}

\section{Introduction}

Deep fakes and misinformation are major societal challenges exacerbated by recent advances in generative models \cite{ldm,imagen,dalle2}.  It has never been more important to understand the origins (or `provenance') of digital images, to enable informed trust decisions to be made on content encountered online.  Emerging standards such as the `Coalition on Content Provenance and Authenticity' (C2PA) \cite{c2pa} embed signed provenance information within image metadata, yet this is easily stripped by attackers or as the image gets shared over the internet (\eg by social media platforms).  Perceptual hashing has been explored for near-duplicate search of trusted databases in order to recover stripped provenance information \cite{nguyen2021,vpn,Zhang2020manip,Bharati2021}.  However, such matches are by definition inexact, and require human-in-the-loop review when matching at scale.  Image watermarking \cite{devi2009,baba2009,weng2019high} provides a potential solution, enabling an identifier to be robustly inserted into an image, which may be used for exact search of a provenance database.  Watermarking may also be used to robustly embed other indicators, such as whether an image has been produced by a generative model.

Watermarking techniques are typically steganographic -- fusing an embedded payload (or `secret') within the  image (`cover') pixels in such a way that the change in the watermarked (stego) image is {\em perceptually invisible} but at the same time {\em recoverable} via a learned secret decoder. This requirement is important for creative work where image quality should be preserved.  The challenge of such processes is to learn both a representative image distribution and the secret embedding process at the same time.  

To this end, we propose RoSteALS, a simple yet effective steganography technique that leverages `free knowledge' from a locked (frozen) autoencoder. Our technical contributions are three-fold:

\textbf{1. Latent steganographic embedding.} We propose a novel method to inject the secret directly into the latent code of a locked autoencoder, enabling robust watermarking with limited training and no content specialization. Our secret encoder is small in size, easy to train, generalize well beyond training data, and can be adapted to the most advanced autoencoders to date.

\textbf{2. Robust secret recovery.} Our approach to secret embedding is shown to withstand severe image perturbations, critical to the use case of persisting identifiers robustly through online redistribution of content.

\textbf{3. Cover-less steganography.} We show that RoSteALS can be easily adapted for cover-less 
 use (\ie where a cover is synthesised on the fly for secret embedding) and for novel text-based steganographic applications. With RoSteALS, we extend the idea of cover-less steganography to an autoencoder's latent space, where our aim is to generate a latent offset that can uniquely represent a secret. Given an autoencoder, the learned offset can be added to any random latent code (generated using an image or text using a diffusion model) of that autoencoder to produce the stego image. 

\section{Related work}

The main goals of steganography techniques are three-fold: maximize the {\em length} of the secret that can be embedded, imperceptibility of the secret in the embedded image, and robustness of the secret decoding against benign and adversarial attacks. The secret can be embedded in the spatial or frequency space of the image using either hand-crafted and learning-based methods. Here we briefly describe the related techniques and place our work in context of others. 

\textbf{Hand-crafted methods} Least significant bit (LSB) embedding was one of the first hand-crafted technique where the secret is embedded in the lowest order bits of each image pixel~\cite{wolfgang1996watermark}, producing images which are perceptually indistinguishable from the original cover image. Ever since there has many techniques designed to embed the secret imperceptibly in the spatial~\cite{taha2022high, ghazanfari2011lsb++} and frequency~\cite{jsteg,navas2008dwt, li2007steganographic, pevny2010using, holub2012designing, holub2014universal,outguess} domain of the image. Even though most of these methods can embed large payloads in the image imperceptibly, they suffer with poor robustness to even minor modifications to the secret image. 

\textbf{Deep learning methods} provide better robustness to noises while maintaining good quality of the generated image~\cite{wan2022comprehensive}. HiDDeN~\cite{zhu2018hidden} was the first end-to-end trained watermarking network that used the encoder-decoder architecture along with a noise layer and adversarial discriminator for robustness and stego-image quality. Given a cover image and a secret of fixed length as inputs, encoder generates a stego image. The decoder takes the noisy stego image and outputs the embedded secret, while the adversarial discriminator compares the stego image and cover image for quality. Using the similar encoder-decoder approach,a variety of architectures were proposed for the improvement of stego image quality and robustness in several later works~\cite{tancik2020stegastamp,zhang2019robust,chang2021neural,subramanian2021end2end, duan2019reversible,wu2018stegnet,meng2018fusion,huang2022image}. These works mostly encode secrets and covers jointly, often with an UNet-like model and skip connections to preserve small details in the cover images \cite{tancik2020stegastamp,duan2019reversible,zhang2019robust}. Different from these techniques, we use a fixed pretrained auto-encoder and train a simple very small network to map the secret to the auto-encoder latent space independent of the cover image. Leveraging pretrained models for steganography has been explored recently -- SSL \cite{fernandez2022watermarking} uses a ResNet50 model pretrained with DINO~\cite{caron2021emerging} self-supervised learning to watermark an image using back-propagation. Here, we demonstrate that the latent space of an autoencoder also provides excellent steganographic capability. Furthermore, SSL requires on-the-fly optimization and cannot operate in a blind setting as it needs a secret key to decode the secret, while our method requires just a single pass forward during inference. 

\textbf{Cover-less steganography} aims to generate a cover image that can uniquely represent a secret, instead of changing a given cover image to embed the secret~\cite{qin2019coverless}. It was first introduced in ~\cite{zhou2015coverless} where the authors indexed a set of cover images to unique hash values that can be transformed to secrets. These hash values can be generated using image properties like pixel value~\cite{zhou2015coverless}, visual bag-of-words~\cite{zhou2016coverless, zhou2018encoding}, or histogram of oriented gradients~\cite{zhou2017coverless}. In ~\cite{volkhonskiy2020steganographic, duan2018coverless}, the authors learned to convert the secret message to GAN latent noise to generate appropriate container images. \cite{liu2022coverless} achieves similar goals via a disentangled structure-texture autoencoder model trained on narrow domains. Other GAN-related works, \cite{yu2021artificial,yu2022responsible}, inject the secret in form of a fingerprint to either training data \cite{yu2021artificial} or model parameters \cite{yu2022responsible} so that the trained models always generate images containing that fingerprint. In contrast to this, given an autoencoder, we learn a unique offset for every secret that can be independently applied to any latent code of that autoencoder. 

\section{Methodology}
\subsection{Leveraging pretrained AutoEncoders}
\begin{figure}[]
    \centering
    \includegraphics[width=1.0\linewidth,trim=0cm 0cm 0cm 0cm,clip]{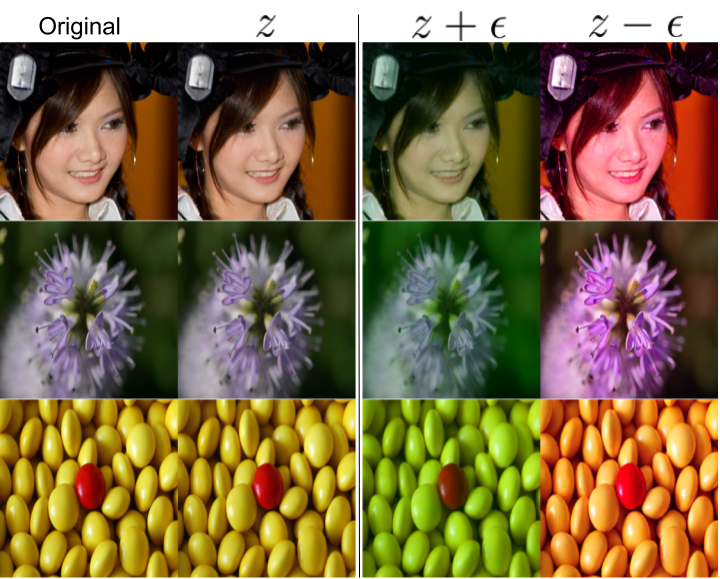}
    \squeezeup
    \caption{Correlation between changes in latent code and pixels of the reconstructed images. Here, the offset $\mathbf{\epsilon}$ is selected as $\mathbf{\epsilon}=k\mathcal{U}\mathbf{\sigma}$}, where $\mathcal{U}$ is a random uniform function, $\mathbf{\sigma}$ is the std.dev. of the VQGAN latent space and noise strength $k=0.2$.
    \squeezeup
    \label{fig:ae_prob}
\end{figure}
To explore whether a pretrained autoencoder has capability for steganography, we experiment with the latent space of the state-of-art autoencoder, VQGAN \cite{esser2020taming}. Given an image $\mathbf{x}$ of size $H\times W\times C$, the VQGAN encoder maps $\mathbf{x}$ into a latent code $\mathbf{z}=E(\mathbf{x}) \in \mathbb{R}^{H'\times W' \times C}$ where $H'$ and $W'$ are typically $4\times$ or $8\times$ smaller than the original resolution. We work on the continuous space of $\mathbf{z}$ therefore the quantization step is performed at the generator side $\bar{\mathbf{x}} = G(\mathbf{z})$ to reconstruct $\mathbf{x}$. We inject different kinds of noise to $\mathbf{z}$ and observe changes on the reconstructed image. We made 2 observations: (i) certain noises cause the same perceptual change on the reconstructed images regardless of its content (\Cref{fig:ae_prob}); and (ii) this embedding space of VQGAN is insensitive to small noise, $G(\mathbf{z}+\mathbf{\epsilon}) \approx \mathbf{x}$ when $\left|\mathbf{\epsilon}\right| < \mathbf{\epsilon}_0$. This is probably due to the subsequent quantization process, however other models such as Kullback–Leibler regulated autoencoders do not have quantization but also exhibit similar behaviors. Regardless of the causes, (ii) inspires a possibility to inject meaningful messages (secrets) directly to the latent code without altering  the image content in a noticeable way, and (i) poses an interesting question of whether the noise can be recovered given the output image. 

\subsection{Steganography with RoSteALS}

\begin{figure*}[t!]
    \centering
    \includegraphics[width=1.0\linewidth,trim=0cm 0cm 0cm 0cm,clip,height=5cm]{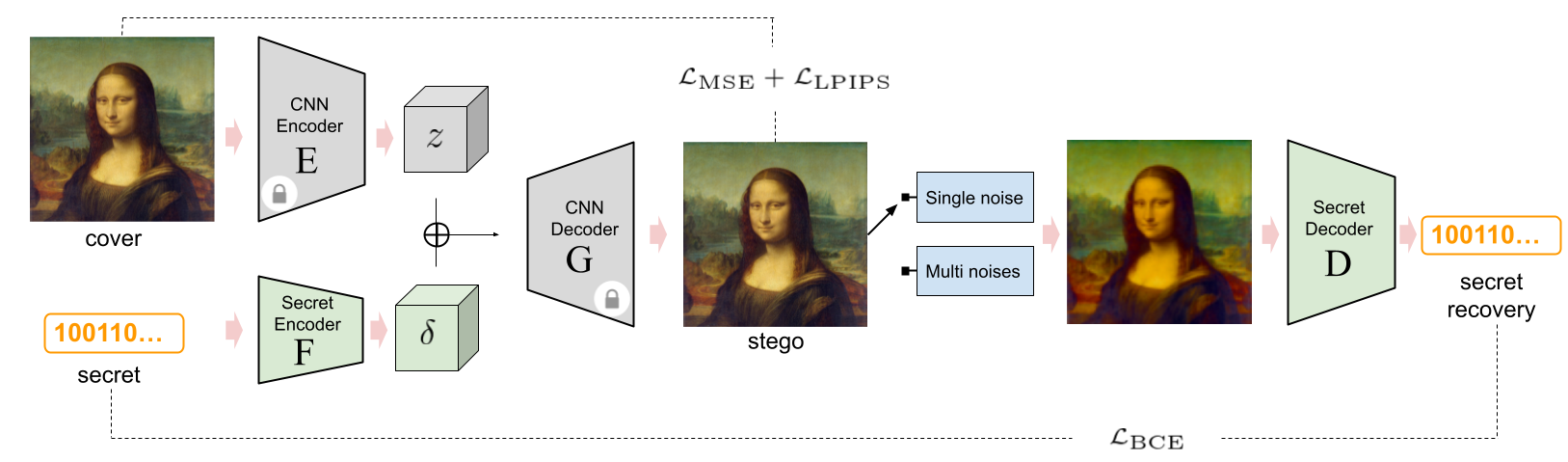}
    \squeezeup
    \caption{Architecture diagram of RoSteALS. The image encoder (E) and decoder (G) are locked during training, only updating the lightweight secret encoder (F) and decoder (D).}
    \squeezeup
    \label{fig:arch}
\end{figure*}

Having prior knowledge of image distribution via the frozen autoencoder \{E, G\}, we aim to learn a secret encoder F to map secret $\mathbf{s} \in \{0,1\}^L$ to the image's latent embedding, where $L$ is the secret length;
\begin{equation}
    \mathbf{\delta} = F(\mathbf{s}) \in \mathbb{R}^{H'\times W' \times C}
\end{equation}

F is a very small network consisting of a fully connected layer followed by SiLU \cite{silu}, then reshaped and upscaled to match the dimension of $\mathbf{z}$, and followed by a final $1\times1$ convolution layer. The output $\mathbf{\delta}$ acts as a small offset to be added to the cover embedding $\mathbf{z}$ (\Cref{fig:arch}). We initialize the weight and bias of the 1x1 convolution layer to 0, following \cite{controlnet}. This is to ensure $\mathbf{\delta}=0$ in the first training iteration, so that the network has exactly the same initial behaviour as the original autoencoder. In total, our secret encoder F has just over 300K parameters for a 100-bit secret length. Interestingly, we empirically verify that conditioning F on both the cover image and secret (\eg $\mathbf{\delta} = F(\mathbf{x},\mathbf{s})$) is not necessary (c.f. \Cref{fig:ae_prob}, see also Sup.Mat.). The stego image is then constructed as $\tilde{\mathbf{x}} = G(\mathbf{z}+\mathbf{\delta})$ and regulated using a combination of pixel and perceptual losses;
\begin{align}
    \mathcal{L}_{MSE} &= \left|\left| \gamma(\tilde{\mathbf{x}}) - \gamma(\mathbf{x}) \right|\right|^2\\
    \mathcal{L}_{\textrm{quality}} &= \mathcal{L}_{LPIPS}(\tilde{\mathbf{x}}, \mathbf{x})+ \alpha \mathcal{L}_{MSE}  \label{eq:quality}
\end{align}
where $\gamma(.)$ is a differentiable non-parametric mapping function from RGB to the more perceptually uniform YUV space; $\mathcal{L}_{LPIPS}$ refers to the LPIPS loss \cite{lpips} commonly used as evaluator for image quality and $\alpha$ is a loss weight constant. 

We use Resnet50 \cite{resnet} as the secret decoder D, replacing the last fully connected layer to output a L-bit secret $\tilde{s}$. We use BCE to compute the bit recovery loss between the predicted and groundtruth secret, $\mathcal{L}_{\textrm{recovery}} = \mathcal{L}_{BCE}(\mathbf{s}, \tilde{\mathbf{s}})$.

The total loss is computed as;
\begin{equation}
\mathcal{L} = \beta \mathcal{L}_{\textrm{quality}} + \mathcal{L}_{\textrm{recovery}}
\label{eq:loss}
\end{equation}
 where loss weight $\beta$ controls the trade-off between stego quality and secret recovery.

\textbf{Training for robustness}. Being robust to noises is a desirable property of both steganography and watermarking. We insert a noise model between the image decoder G and the secret decoder D. We use 14 noise sources from ImageNet-C \cite{imagenetc}, a rich and diverse library commonly used for evaluating model robustness. These noises can be split into 3 groups: {\em differentiable} including most additive and linear noises (\eg brightness, saturation, contrast, ...), {\em approximatable with differentiable transforms} (\eg jpeg compression \cite{shin2017jpeg}), {\em non-differentiable} (\eg spatter). To ensure gradient can be flown back to the preceding layers, we convert any non-differentiable noise $n(.)$  to additive noise via $n(\mathbf{x})=\mathbf{x} + [n(\mathbf{x})-\mathbf{x}]$ where $[.]$ is treated as an additive constant (gradient disconnected from computation graph). Additionally, we apply multiple common noises (\eg contrast, brightness, jpeg compression) concurrently at a certain rate ($p=0.5$), along with random individual noises. This enables the secret decoder to be trained under various data augmentations and  feedback signal can be backpropagated to update the secret encoder.

\begin{figure*}[t!]
    \centering
    \includegraphics[width=1.0\linewidth,trim=0cm 0cm 0cm 0cm,clip,height=10cm]{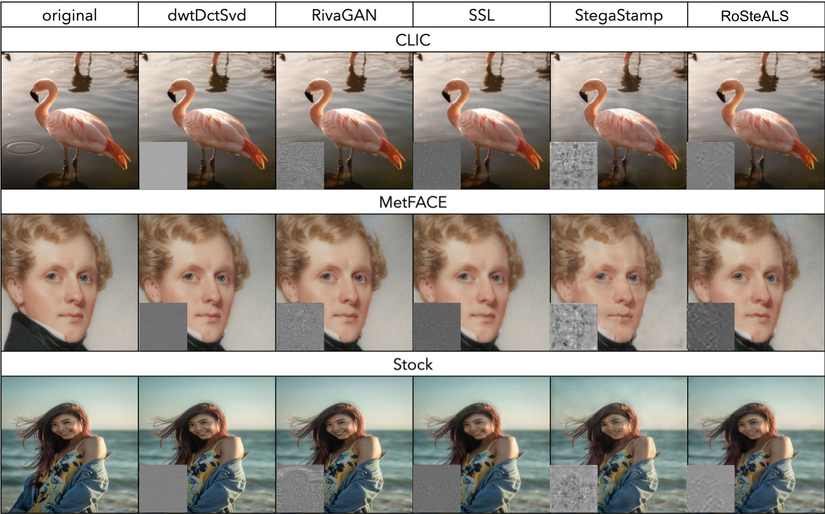}
    \squeezeup
    \caption{Qualitative examples of watermarked images from different techniques. The secret is the binary form of word string ``some secrets". In each case, except StegaStamp, the watermarked images look similar to the original images with no perceptual artifacts. Notice in the case of StegaStamp, there are visible artifacts in the entire image (second row) and on the woman's forehead in the third row.}
    \squeezeup
    \label{fig:qualitative}
\end{figure*}

\section{Experiments}
\subsection{Datasets, training details and metrics}
\label{sec:data}
\textbf{Datasets}. We train RoSteALS using 100K images and validate on 1K images from the MIRFlickR dataset \cite{mirflick}. We evaluate on 3 different benchmarks - CLIC \cite{clic2020}, MetFace \cite{metface} and Stock1K - our own collection of 1K images on Stock \footnote{\url{https://github.com/TuBui/RoSteALS}}. CLIC \cite{clic2020} contains a mix of 530 high quality mobile and professional photographs often used to evaluate image compression and rate-distortion techniques. MetFace \cite{metface} is a narrow-domain collection of 1336 human faces extracted from artwork. Stock1K has 1000 high quality multimedia images, including photographs, vectorarts, sketchs and graphic designs. We choose these 3 datasets for diversity in content, style and domain (\Cref{fig:qualitative}). To make the benchmarks more challenging, we apply a random ImageNet-C perturbation on every output stego.

\textbf{Training details}. Images are randomly cropped and resized to $256 \times 256$ during training. At test time, the whole images are resized to the specified resolution. We use the pretrained $256 \times 256$ VQ-f4 model of VQGAN \cite{esser2020taming} for most experiments, but also explore the $512 \times 512$ KL-f8 version (\Cref{sec:coverless}). Unless otherwise stated, the secret size is fixed at L=100 bits and uniformly sampled at random during training and testing. We use the AdamW optimizer and learning rate 8e-5 using the Pytorch library. Training is terminated when the moving average of validation loss stops improving. We set $\alpha=1.5$ in \Cref{eq:quality} and the quality loss weight $\beta$ in \Cref{eq:loss} is determined dynamically. Specifically, we find that it is critical to prioritize $\mathcal{L}_{\textrm{recovery}}$ at the early training phase, since image quality is guaranteed in the first iteration through the locked autoencoder \{E,G\} while the secret encoder/decoder \{F, D\} must be learnt from scratch. We therefore start with a fixed image batch, set a low value of $\beta =0.1$ and train our network to prioritize prediction of the random secret $s$. We unlock the full training image set after bit accuracy reaches a certain threshold $t_1$, and linearly increase $\beta$ till $\beta_{\textrm{max}}$ and turn on the noise model after a higher bit accuracy threshold $t_2$ is met. We choose $t_1=90\%, t_2=98\%$ and $\beta_{\textrm{max}}=10$ and do not tune it extensively. More details are in Sup.Mat.      

\textbf{Metrics}. To evaluate stego quality versus the original cover image, we employ Peak Signal To Noise Ratio (PSNR), Structural Similarity Index Measure (SSIM), perceptual similarity score (LPIPS \cite{lpips}) and Single Image Frechet Inception Distance (SIFID \cite{sifid}). For secret recovery evaluation,  we report standard bit accuracy ({\em Bit acc.}) and {\em Bit acc. (ECC)} (bit accuracy with cyclic error correction code using BCH \cite{bch}, following the settings in \cite{volkhonskiy2020steganographic}). We also compute accuracy at word level, {\em Word acc.}, where a match is considered successful if the predicted secret differs from the corresponding groundtruth in less than 20\% bits. Additionally, we report bit accuracy on clean stego for reference. Note that apart from {\em Bit acc. (clean)}, three other metrics work on noised data. 

\subsection{Baseline comparison}

\begin{figure*}[t!]
    \centering
    \includegraphics[width=1.0\linewidth,trim=0cm 0cm 0cm 0cm,clip]{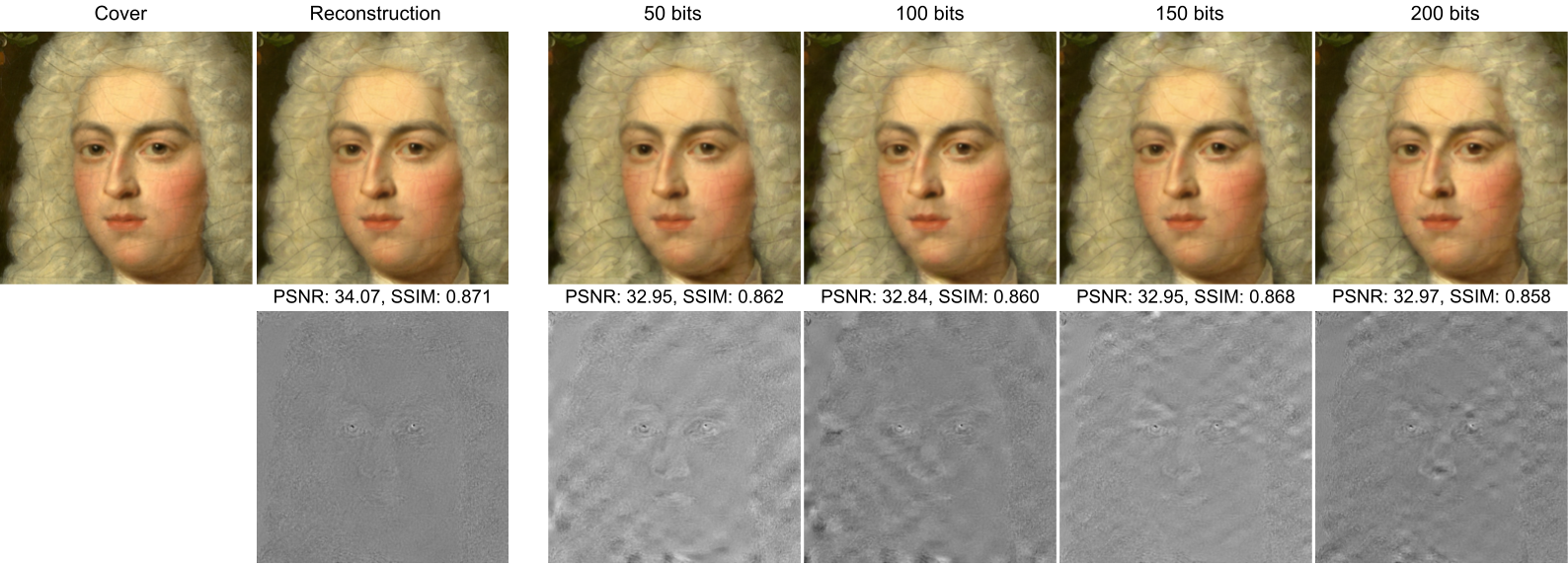}
    \squeezeup
    \caption{Change in image quality as secret length increases. Residual images are scaled to [0,255] range for visualization purpose.}
    \squeezeup
    \label{fig:bitlen_eg}
\end{figure*}

We compare RoSteALS with the following baselines:
\begin{itemize}
    \item \textbf{dwtDctSvd}~\cite{navas2008dwt}: a state-of-art handcrafted frequency-based method that performs secret embedding in the U frame of the YUV space after a sequence of discrete wavelet transform, discrete cosine transform and singular value decomposition. We do not report results for other popular handcrafted methods such as LSB, Jsteg \cite{jsteg} and OutGuess \cite{outguess} because these are not designed to handle noises other than JPEG compression.

    \item \textbf{StegaStamp}~\cite{tancik2020stegastamp}: a CNN-based method with state-of-art robustness performance.

    \item \textbf{RivaGAN}~\cite{zhang2019robust}: a GAN-based method leveraging attention mechanism and adversarial training.

    \item \textbf{SSL}~\cite{fernandez2022watermarking}: an on-the-fly optimization method leveraging pretrained representation learning models.
    
\end{itemize}
\begin{figure}[t!]
    \centering
    \includegraphics[width=1.0\linewidth,trim=0cm 0cm 0cm 0cm,clip]{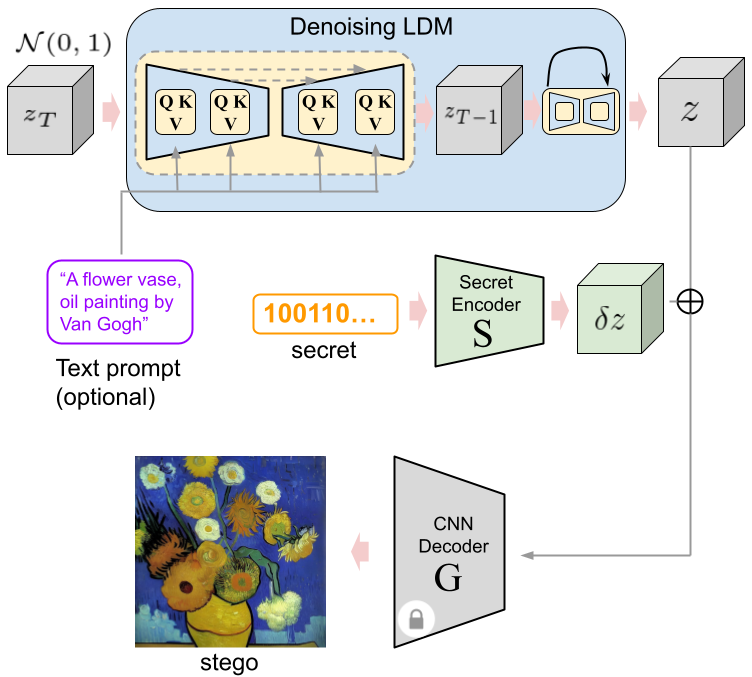}
    \squeezeup
    \caption{Text-based and cover less steganography.}
    \squeezeup
    \label{fig:coverless_arch}
\end{figure}
For StegaStamp, we use the officially released model pretrained on MIRFlickR for L=100 bits as we were not able to reproduce the reported results. Models for other methods are trained with the same settings as RoSteALS using their public code.

\begin{table*}[t!]
\centering
\definecolor{Gray}{gray}{0.9}
\small
\resizebox{\linewidth}{!}{
\begin{tabular}{l|cccc|cccc}
\toprule
& \multicolumn{4}{c|}{Image quality} & \multicolumn{4}{c}{Secret recovery} \\
Method     & PSNR $\Uparrow$       & SSIM $\Uparrow$ & LPIPS $\Downarrow$     & SIFID $\Downarrow$ & Bit acc. (clean) $\Uparrow$ & Bit acc. $\Uparrow$& Bit acc. (ECC) $\Uparrow$ & Word acc. $\Uparrow$\\
\midrule 
\multicolumn{9}{c}{\textbf{CLIC}}                                                                                                \\
\rowcolor{Gray} RoSteALS     & 32.68 $\pm$ 1.75 & 0.88 $\pm$ 0.06 & 0.04 $\pm$ 0.02 & 0.04 $\pm$ 0.02 & \textbf{1.00} & \textbf{0.94 $\pm$ 0.07} & \textbf{1.00} & \textbf{0.93} \\
VQGAN   & 33.90 $\pm$ 14.47 & 0.90 $\pm$ 0.06 & 0.03 $\pm$ 0.02 & 0.02 $\pm$ 0.02 & N/A & N/A & N/A & N/A \\
StegaStamp \cite{tancik2020stegastamp} & 31.26 $\pm$ 0.85 & 0.91 $\pm$ 0.03 & 0.09 $\pm$ 0.03 & 0.23 $\pm$ 0.13 & \textbf{1.00} & 0.88 $\pm$ 0.13 & 0.48 $\pm$ 0.50 & 0.74 \\

SSL \cite{fernandez2022watermarking}       & \textbf{41.84 $\pm$ 0.10}    &   \textbf{0.98 $\pm$ 0.01}  &   0.02 $\pm$ 0.01  &   \textbf{0.01 $\pm$ 0.02}   &       0.99 $\pm$ 0.03          &               0.62 $\pm$ 0.14   &  0.03 $\pm$ 0.17   &    0.13         \\
RivaGAN \cite{zhang2019robust}   &  40.32 $\pm$ 0.15   &   \textbf{0.98 $\pm$ 0.01}  &   0.02 $\pm$ 0.02  &   0.07 $\pm$ 0.06        &     0.98 $\pm$ 0.03            &        0.77 $\pm$ 0.16          &  0.22 $\pm$ 0.41  &     0.45         \\
dwtDctSvd  \cite{navas2008dwt} & 38.96 $\pm$ 1.41    &   0.97 $\pm$ 0.02  &\textbf{0.01 $\pm$ 0.01} &   0.02 $\pm$ 0.02  &      \textbf{1.00}           &               0.61 $\pm$ 0.20   &  0.16 $\pm$ 0.34  &   0.21           \\
\midrule
\multicolumn{9}{c}{\textbf{MetFACE}}                                                                                                \\
\rowcolor{Gray} RoSteALS     & 34.46 $\pm$ 1.91 & 0.89 $\pm$ 0.07 & 0.04 $\pm$ 0.02 & \textbf{0.01 $\pm$ 0.02} & \textbf{1.00} & \textbf{0.94 $\pm$ 0.08} & \textbf{1.00} & \textbf{0.91}   \\
VQGAN & 35.98 $\pm$ 2.45 & 0.90 $\pm$ 0.07 & 0.02 $\pm$ 0.02 & \textbf{0.01 $\pm$ 0.02} & N/A & N/A & N/A & N/A   \\
StegaStamp \cite{tancik2020stegastamp}&  32.01 $\pm$ 0.77 & 0.92 $\pm$ 0.02 & 0.13 $\pm$ 0.03 & 0.22 $\pm$ 0.15 & 1.00 & 0.86 $\pm$ 0.14 & 0.47 $\pm$ 0.50 & 0.68    \\
SSL \cite{fernandez2022watermarking}       &  \textbf{41.77 $\pm$ 0.12}   &   \textbf{0.98 $\pm$ 0.01}  &   0.04 $\pm$ 0.02  &   0.04 $\pm$ 0.05  &       \textbf{1.00}           &  0.63 $\pm$ 0.16                &   0.08 $\pm$ 0.27  &   0.19          \\
RivaGAN \cite{zhang2019robust}   &  40.27 $\pm$ 0.09   &   0.97 $\pm$ 0.01  &   0.06 $\pm$ 0.03  &   0.16 $\pm$ 0.12  &       0.99 $\pm$ 0.01           &             0.78 $\pm$ 0.17     &  0.28 $\pm$ 0.44   &  0.47           \\
dwtDctSvd  \cite{navas2008dwt} &    40.86 $\pm$ 2.48 &   \textbf{0.98 $\pm$ 0.01}  &   \textbf{0.02 $\pm$ 0.01}  &   0.03 $\pm$ 0.02  &      \textbf{1.00}            &              0.63 $\pm$ 0.23    &   0.22 $\pm$ 0.38   &  0.26       \\
\midrule
\multicolumn{9}{c}{\textbf{Stock1K}}                                                                                                \\
\rowcolor{Gray} RoSteALS     &  33.27 $\pm$ 2.32 & 0.89 $\pm$ 0.08 & 0.03 $\pm$ 0.02 & 0.05 $\pm$ 0.06 & \textbf{1.00} & \textbf{0.92 $\pm$ 0.10} & \textbf{1.00} & \textbf{0.864}   \\
VQGAN     &  34.44 $\pm$ 2.71 & 0.91 $\pm$ 0.07 & 0.02 $\pm$ 0.02 & 0.03 $\pm$ 0.06 & N/A & N/A & N/A & N/A \\
StegaStamp \cite{tancik2020stegastamp}&   31.42 $\pm$ 0.95 & 0.92 $\pm$ 0.03 & 0.08 $\pm$ 0.04 & 0.20 $\pm$ 0.14 & \textbf{1.00} & 0.87 $\pm$ 0.13 & 0.48 $\pm$ 0.50 & 0.72    \\
SSL \cite{fernandez2022watermarking}       &    \textbf{42.07 $\pm$ 0.50} &   \textbf{0.99 $\pm$ 0.01}  &   0.02 $\pm$ 0.02  &   \textbf{0.01 $\pm$ 0.02}  &          0.95 $\pm$ 0.09        &            0.59 $\pm$ 0.12      &   0.02 $\pm$ 0.13  &     0.09         \\
RivaGAN  \cite{zhang2019robust}  &    40.49 $\pm$ 0.45 &   0.98 $\pm$ 0.01  &   0.02 $\pm$ 0.02  &   0.05 $\pm$ 0.06  &        0.93 $\pm$ 0.09          &   0.72 $\pm$ 0.16               &   0.13 $\pm$ 0.33   & 0.31          \\
dwtDctSvd  \cite{navas2008dwt} &   39.76 $\pm$ 2.41  &   0.98 $\pm$ 0.02  &   \textbf{0.01 $\pm$ 0.01}  &   0.02 $\pm$ 0.02  &        0.95 $\pm$ 0.13          &          0.60 $\pm$ 0.19        &  0.17 $\pm$ 0.33    &   0.18         \\

\bottomrule
\end{tabular}}
\squeezeupsmall
\caption{Comparison of RoSteALS and baseline methods for image quality and secret recovery performance. Values for each metric are reported as mean and standard deviation. VQGAN is the autoencoder backbone of RoSteALS. Best results are in bold.}
\squeezeupsmall
\label{tab:main}
\end{table*}

\Cref{tab:main} shows image quality and secret recovery performance of all methods. We also include quality evaluation for RoSteALS's autoencoder, VQGAN \cite{esser2020taming}, for reference.  There is not a clear winner in stego image quality metric. SSL \cite{fernandez2022watermarking} achieves the best PSNR and SSIM scores on all benchmarks while dwtDctSvd is most effective on LPIPS metric. On SIFID, SSL outperforms the rest on CLIC and Stock1K, but RoSteALS is the winner for MetFACE. This implies that the pretrained feature extraction model in SSL does not generalize as good as the autoencoder in RoSteALS. StegaStamp performs the worst in general, although still slightly outperforms RoSteALS on SSIM (see \Cref{sec:conclusion} for explanation). We note that RoSteALS performance is capped by its autoencoder backbone, trading just 1-2 points in each metric for secret embedding capacity and could be improved further with better backbone in future. \Cref{fig:qualitative} shows example of stego images in each benchmark. 

\begin{figure}[t!]
    \centering
    \includegraphics[width=1.0\linewidth,trim=0cm 0cm 0cm 0cm,clip]{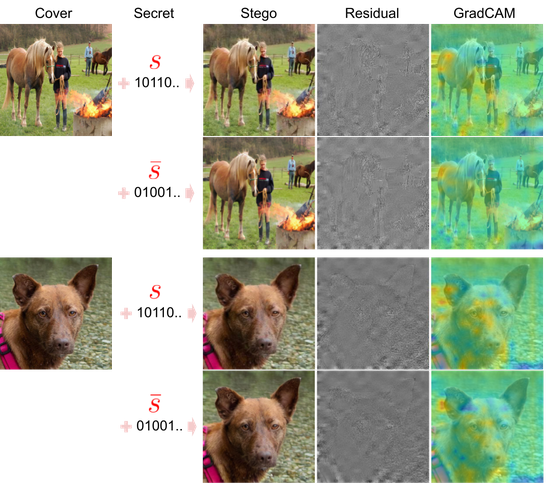}
    \squeezeup
    \caption{Studies into how the same secret and its inverted form are hidden in two different cover images. Residual images are scaled to range [0,255] for visualization purpose. Note the similarity in GradCAM heatmap between rows 1\&3, and rows 2\&4.}
    \squeezeup
    \label{fig:gradcam}
\end{figure}

In term of secret recovery, all methods achieves near perfection score on clean data. RoSteALS outperforms the rest in all metrics and StegaStamp is the followup winner, while the handcrafted method and SSL have the lowest performance. The diversity in {\em Bit acc. (ECC)} shows a side effect of error correction - greatly improving performance for slightly corrupted data (as in case of RoSteALS) but also corrupting more if the data has damage beyond a certain threshold (other baselines). We attribute the performance of RoSteALS to its capability of separating the secret embedding and robustness learning tasks from image distribution learning task, resulting in perfect performance in both clean and noise data (with ECC).   

\subsection{Robustness}
\label{sec:robust}
\begin{figure*}[t!]
    \centering
    \includegraphics[width=1.0\linewidth,trim=0cm 0cm 0cm 0cm,clip]{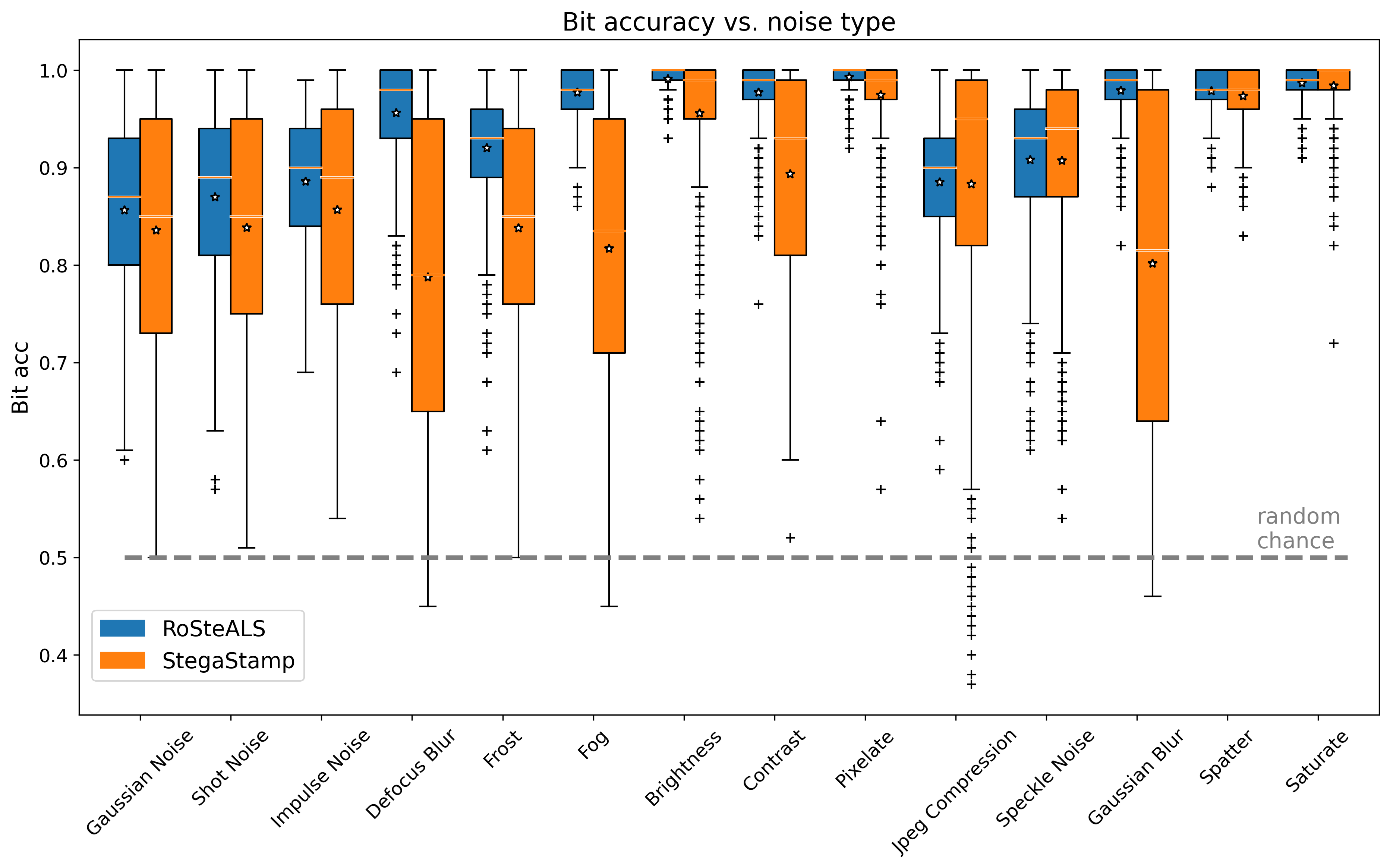}
    \squeezeup
    \caption{Effects of different perturbation sources to bit accuracy performance of RoSteALS and StegaStamp. Random chance is 50\%.}
    \label{fig:noise}
    \squeezeup
\end{figure*}

\Cref{fig:gradcam} depicts the changes on stego image when two different secrets (inversion of each other) are embedded, as well as the effects of embedding the same secret on two different images. The residual the GradCAM heatmaps \cite{gradcam} indicates that the secret is embedded across the entire image. Additionally, it can be seen that the secret embedding is not dependent on image content (c.f. \Cref{fig:ae_prob}). This properties coupled with the generalized VQGAN autoencoder benefit RoSteALS`s generalization on new domains unseen during training.  

\begin{figure}[t!]
    \centering
    \includegraphics[width=1.0\linewidth,trim=0cm 0cm 0cm 0cm,clip]{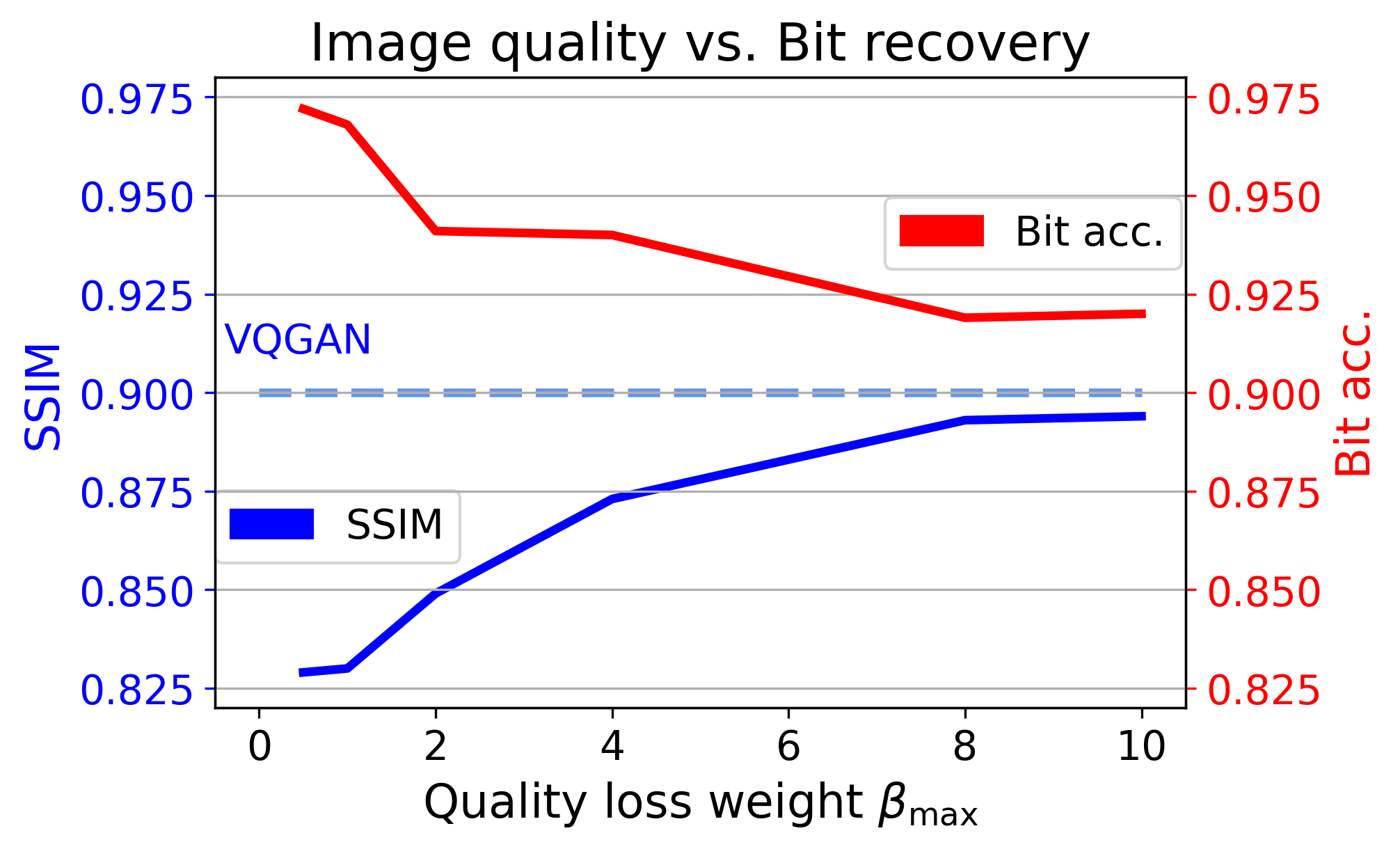}
    \squeezeup
    \caption{Stego quality and secret recovery trade-off, tested on the Stock1K benchmark. Performance of the VQGAN backbone serves as an upper limit for SSIM.}
    \squeezeup
    \label{fig:loss_weight}
\end{figure}

\begin{table}[t!]
\centering
\begin{tabular}{l|cccc}
\toprule
Secret length (bits) & 50 & 100 & 150 & 200 \\
 \midrule
PSNR & 32.81 & 32.69 & 32.85 & 32.89 \\
SSIM & 0.89 & 0.88 & 0.88 & 0.88 \\
Bit acc. & 0.97 & 0.94 & 0.87 & 0.84 \\
Train time (epochs) & 17 & 18 & 21 & 30\\
\bottomrule
\end{tabular}
\squeezeupsmall
\caption{Image quality, secret recovery performance and training speed versus secret length on the CLIC dataset. All stego images are subjected to random noises.}
\squeezeup
\label{tab:bitlen}
\end{table}

We analyze the secret recovery performance of RoSteALS and its closest competitor against individual noise sources in \Cref{fig:noise}. RoSteALS is more robust and stable than StegaStamp against all perturbations, especially on blurring (Gaussian, defocus) and heavy image enhancement effects (frost, fog). Highest performance is acquired on simple linear noises such as brightness, contrast and saturation as well as pixelate, for both methods. On the other hand, heavy jpeg compression and those that completely wipe values of certain pixels (shot, impulse and speckle noises) are the most challenging.

\begin{figure}[t!]
    \centering
    \begin{tabular}{c}
         \includegraphics[width=1.0\linewidth,trim=0cm 0cm 0cm 0cm,clip]{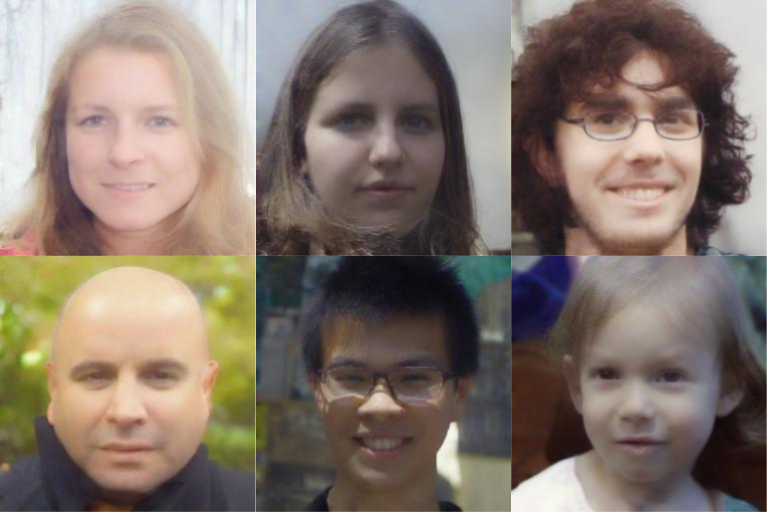} \\
         \includegraphics[width=1.0\linewidth,trim=0cm 0cm 0cm 0cm,clip]{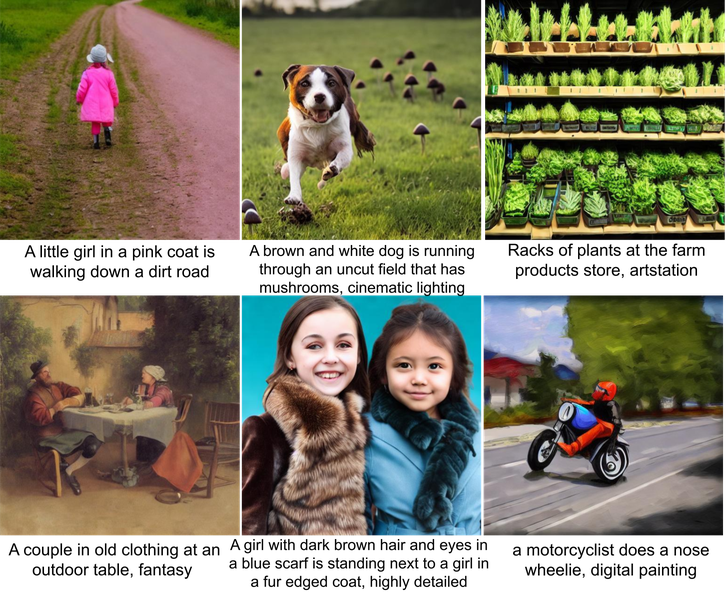}
    \end{tabular}
    \squeezeupsmall
    \caption{Random stego images are generated using our unconditional LDM-RoSteALS (top) and text conditioned LDM-RoSteALS (bottom), covering random secrets. All has bit acc. $>99\%$. The LDM model in top row is trained on FFHQ and the LDM in bottom row is the pretrained Stable Diffusion v1.5 model. }
    \squeezeup
    \label{fig:coverless}
\end{figure}

\begin{table}[]
\centering
\begin{tabular}{l|cccc}
\toprule
Train volume ($\times10^3$) & 10 & 40 & 80 & 100 \\
 \midrule
PSNR & 35.07 & 34.51 & 34.36 & 34.46 \\
SSIM & 0.89 & 0.90 & 0.90 & 0.89 \\
Bit acc. & 0.91 & 0.91 & 0.93 & 0.94 \\
\bottomrule
\end{tabular}
\squeezeupsmall
\caption{Image quality and secret recovery performance versus amount of training data on the MetFace dataset. All stego images are subjected to random noises.}
\squeezeupsmall
\label{tab:vol}
\end{table}

\subsection{Dependencies and the quality/recovery tradeoff}
We study the effect of secret length on RoSteALS performance and training speed. \Cref{tab:bitlen} demonstrates RoSteALS abilities to retain the same image quality as secret length increases, at the cost of decreasing secret recovery performance. Increasing secret length from 50 to 200 bits causes 13\% loss in bit accuracy. 
\Cref{fig:bitlen_eg} shows an example of stego images produced by RoSteALS models trained on different secret length. Similarly, \Cref{tab:vol} shows that image quality is not affected by the training volume either, thanks to the frozen autoencoder. Secret recovery performance is affected due to overfitting at the secret decoder (bit accuracy increases by 3\% at 10x increase in the training volume).

The training time also increases with longer secret length (\Cref{tab:bitlen}). However, even at L=200 it only takes 30 epochs for the training to converge, as RoSteALS sole goal is to learn the secret encoding and decoding modules. For reference, training on SSL requires 100 passes through every images and RivaGAN requires 300 epochs to converge.  

There is a trade-off between stego quality and secret recovery. \Cref{fig:loss_weight} shows that it is possible to control this trade-off in RoSteALS using the loss weight $\beta_{\textrm{max}}$, which dictates the maximum value of $\beta$ in \Cref{eq:loss} (see \Cref{sec:data}). Increasing $\beta_{\textrm{max}}$ by 10 times boosts the SSIM score by 7points while reducing the {\em Bit acc.} score by 5\%. It is worth noting that the trade-off can not go further beyond $\beta_{\textrm{max}}=10$ as the image quality is constrained by the autoencoder performance.   

\subsection{Text-based and Cover-less stega}
\label{sec:coverless}

\begin{table}[t!]
    \centering
    \begin{tabular}{c|cccc}
    \toprule
        Method & Bit acc. (clean) & Bit acc. & Word acc.  \\
         \midrule
       Cover-less & 0.997& 0.924 & 0.875  \\
       Text-based &  0.992 & 0.904 & 0.844 \\
       \bottomrule
    \end{tabular}
    \squeezeupsmall
    \caption{Secret recovery performance for unconditional (cover-less) and text-based conditional LDM-RoSteALS.}
    \squeezeupsmall
    \label{tab:coverless}
\end{table}

RoSteALS's design of injecting the secret signal directly to the already well-defined latent space enables novel applications in cover-less and text-based steganography. In theory, we can remove the image encoder E and inject the secret embedding to a random point in $\mathbf{z}$ space to create a stego image without a cover. However, the distribution of latent code in VQGAN is rather complex -- picking a random $\mathbf{z}$ often lead to low quality output. Instead we can learn a mapping from a simpler distribution (\eg Gaussian for cover-less stega) or from distribution of a different modality (\eg text). We propose to learn such mapping via the denoising process of latent diffusion (LDM) \cite{ldm}, as shown in \Cref{fig:coverless_arch}. Training the LDM is separate from RoSteALS, as long as they share the same autoencoder \{E,G\}. \Cref{fig:coverless} shows examples of cover-less and text-based stego generation. For the cover-less case, the LDM model is trained on FFHQ and we simply reuse our RoSteALS model above. For text-based stega, we use a pretrained Stable Diffusion model and train a new RoSteALS variant with the KL-f8 autoencoder backbone employed in Stable Diffusion. We use the same MIRFlickR training dataset, resize all input images to $512\times512$ for compatibility with the KL-f8 model and set the secret length L=64bit for simplicity.

\begin{figure}[t!]
    \centering
    \includegraphics[width=1.0\linewidth,trim=0cm 0cm 0cm 0cm,clip]{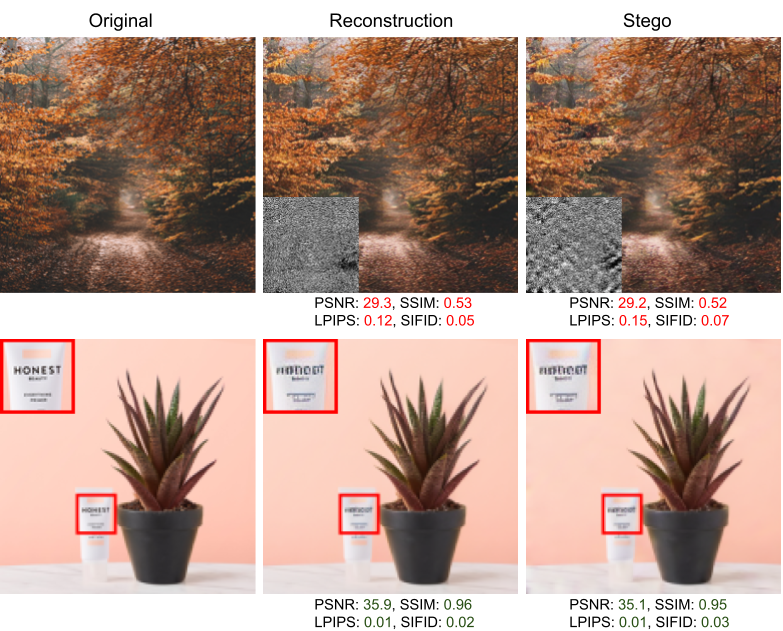}
    \squeezeupsmall
    \caption{Failure cases are caused by VQGAN failing to reconstruct exact details of (top) clutter objects or (bottom) small text. In the top row, the change is hard to perceive but caught by quality metrics (insets are residual scaled to [0,255] for visualization purpose). In the bottom row, artifacts are only local hence still yields high quality scores, but perceptually visible. }
    \squeezeupsmall
    \label{fig:fail}
\end{figure}

To evaluate secret recovery performance, we sample 1000 stego images for each LDM-RoSteALS models using randomly generated secrets. For text-based model, we also leverage 1000 random captions from the Flick8K dataset \cite{flick8k} as the conditioning signals. To enhance the image generation quality of Stable Diffusion, we append a random {\em catching} words (\eg ``highly detailed", ``cinematic lighting", ``artstation", ``sharp focus") at the end of each caption. \Cref{tab:coverless} shows that RoSteALS performs well in both cases, despite never `seeing' a latent code generated by the LDM model during training.
\section{Conclusion}
\label{sec:conclusion}
We demonstrate that the latent space of a pretrained autoencoder can be leveraged to hide data, opening up a new direction for steganography and digital wartermarking applications. We propose a novel secret hiding technique that offers several advantages: small model size, modular design, producing state-of-art secret recovery capability and comparable image quality versus the autoencoder. Our method, RoSteALS, can be easily adapted for novel applications such as cover-less and text-based steganography.  

\textbf{Limitations} - RoSteALS relies on a pretrained autoencoder for image encoding and generation, thus inherits the same drawbacks of this model in preserving small details during image generation. We identify two failure cases of RoSteALS in \Cref{fig:fail} involving cluttered objects that cause lots of tiny spatial shifts during image reconstruction (perceptual invisible but affecting quality measurement metrics, especially SSIM), or the challenge of reconstructing small text or face (not affecting quality metrics but could be perceptually visible). Future work could leverage more powerful autoencoders, or novel ways to carefully finetune the autoencoder that benefit steganography. 

\noindent\textbf{Acknowledgment} This work was supported in part by DECaDE under EPSRC grant EP/T022485/1.
{\small
\bibliographystyle{ieee_fullname}
\bibliography{egbib}
}
\clearpage
\begin{appendices}
\section{Training details}
RoSteALS is easy to train as long as it priorities the secret recovery loss at the early training phase. In \Cref{sec:data} we propose a training method to overcome the complexity of the cover image domain (\eg MIRFlickR is harder to train than FFHQ), the gradient flow between pretrained and learning-from-scratch modules, the challenges of large secret size, and the difficulties for the secret decoder to `learn' perceptually invisible secret signals present in already high-quality images but corrupted with various perturbations. We adopt curriculum learning in our training schedule, starting from a fixed minibatch of cover images without noise corruptions, before unleashing the full training database and eventually enabling perturbations and linear loss weight ramping.

\begin{figure*}[t!]
    \centering
    \includegraphics[width=1.0\linewidth,trim=0cm 0cm 0cm 0cm,clip]{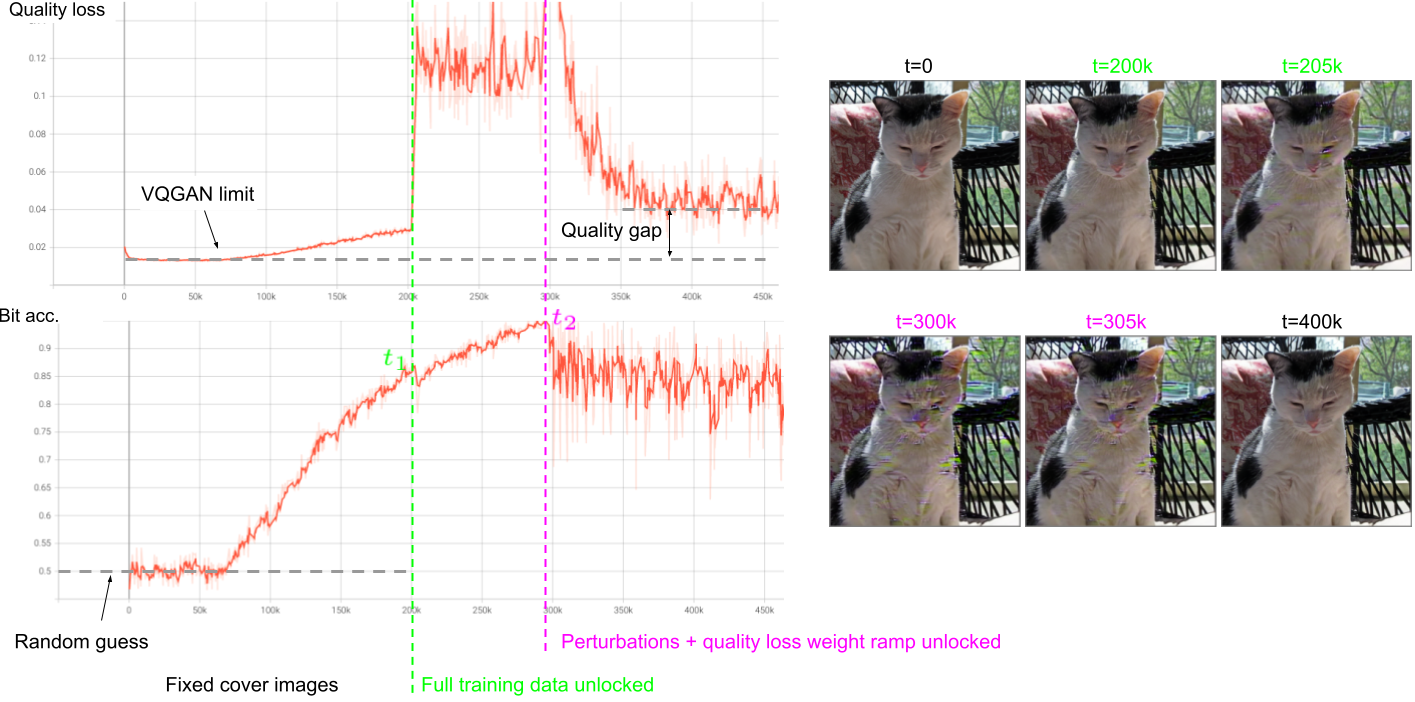}
    \caption{(Left) Quality loss and secret recovery curves for the first 18 epochs when training the 200bit-secret RoSteALS model on MIRFlickR with mini-batch size set to 4. (Right) Evolution of the stego image at different training stages.}
    \label{fig:train}
\end{figure*}

A successful training pipeline should be similar to \Cref{fig:train}. We only experiment with few ($t_1,t_2,\beta_{\textrm{max}}$) tuples and settle with ($t_1=0.90,t_2=0.98,\beta_{\textrm{max}}=10.0$), therefore believe that performance could potentially be improved further with more careful parameter tuning.   

\section{Architecture details}
RoSteALS has a very light-weight secret encoder and can be constructed using just 1 line of code using the Pytorch library. For example, for a 100-bit secret encoder:
\begin{verbatim}
secret_encoder = nn.Sequential(
  nn.Linear(100,32*32*3), nn.SiLU(),
  Lambda(lambda x: x.view(-1,3,32,32)),
  nn.Upsample((2,2)),
  nn.Conv2d(3,3,3,padding=1) 
)
\end{verbatim}
We experimented with more advanced architectures and found no clear benefits over this simple module. 

\begin{figure*}[t!]
    \centering
    \includegraphics[width=1.0\linewidth,trim=0cm 0cm 0cm 0cm,clip]{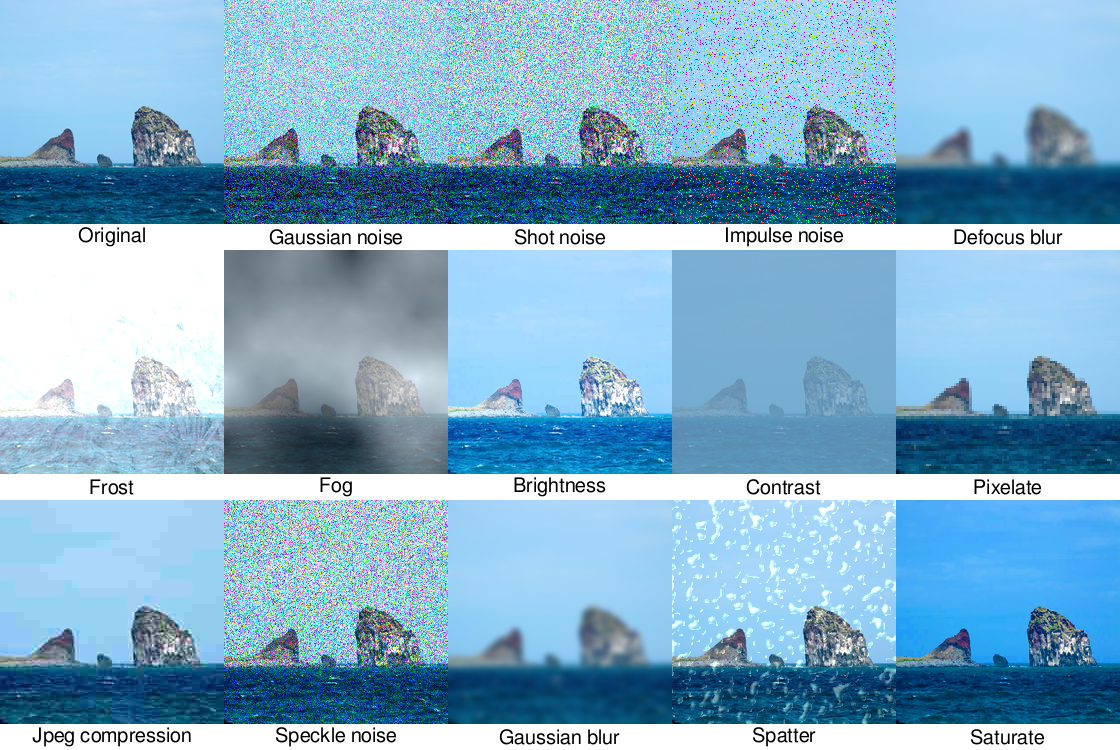}
    \caption{ImageNet-C perturbations on an example image, noise strength is set to 3 (out of 5).}
    \label{fig:netc}
\end{figure*}

\section{Joint cover-secret conditioning}
Existing works often model secrets and covers jointly, arguing the secret embedding should depend on the cover image for optimal stego quality. We observe that is not the case for RoSteALS, as shown in \Cref{fig:ae_prob,fig:gradcam} and discussed in \Cref{sec:robust}. Here, we implemented a RoSteALS alternative with the secret encoder E taking both the secret and cover as inputs. Specifically, the cover image is first blurred and downsampled to $H'\times W'\times C$, retaining only low frequency components. We then concatenate it with the upsampled secret embedding and passing to a sequence of convolution layers with SiLU activation. The weights of the last convolution layer is initialized with 0, in the same way as the proposed RoSteALS.   

\begin{table}[t!]
    \centering
    \small
    \resizebox{\linewidth}{!}{
    \begin{tabular}{c|cccc}
    \toprule
         & PSNR & LPIPS & Bit acc. & Word acc. \\
    \midrule
    \multicolumn{5}{c}{\textbf{CLIC}} \\
        Proposed &  \textbf{32.68 $\pm$ 1.75} & \textbf{0.04 $\pm$ 0.02} & 0.94 $\pm$ 0.07 & \textbf{0.93} \\ 
        Joint cond.  & 32.45+-1.67 & 0.05+-0.02 & 0.94+-0.09 & 0.92  \\
    \midrule
        \multicolumn{5}{c}{\textbf{MetFace}} \\
        Proposed &  \textbf{34.46 $\pm$ 1.91} & 0.04 $\pm$ 0.02 &\textbf{0.94 $\pm$ 0.08} & \textbf{0.91} \\ 
        Joint cond.  & 33.99+-1.81 & 0.04+-0.02 & 0.93+-0.09 & 0.90 \\
    \midrule
    \multicolumn{5}{c}{\textbf{Stock1K}} \\
        Proposed &  \textbf{33.27 $\pm$ 2.32} & \textbf{0.03 $\pm$ 0.02} & 0.92 $\pm$ 0.10 & 0.86 \\ 
        Joint cond.  & 33.00+-2.18 & 0.04+-0.02 & 0.92+-0.11 & 0.86  \\

    \bottomrule
    \end{tabular}
    }
    \caption{Joint cover-secret conditioning provides no benefit in RoSteALS design.}
    \label{tab:joint}
\end{table}

\Cref{tab:joint} shows the performance of this joint conditioning configuration, which is equal or slightly worse than the proposed approach in all metrics. 

\begin{figure*}[t!]
    \centering
    \includegraphics[width=1.0\linewidth,trim=0cm 0cm 0cm 0cm,clip]{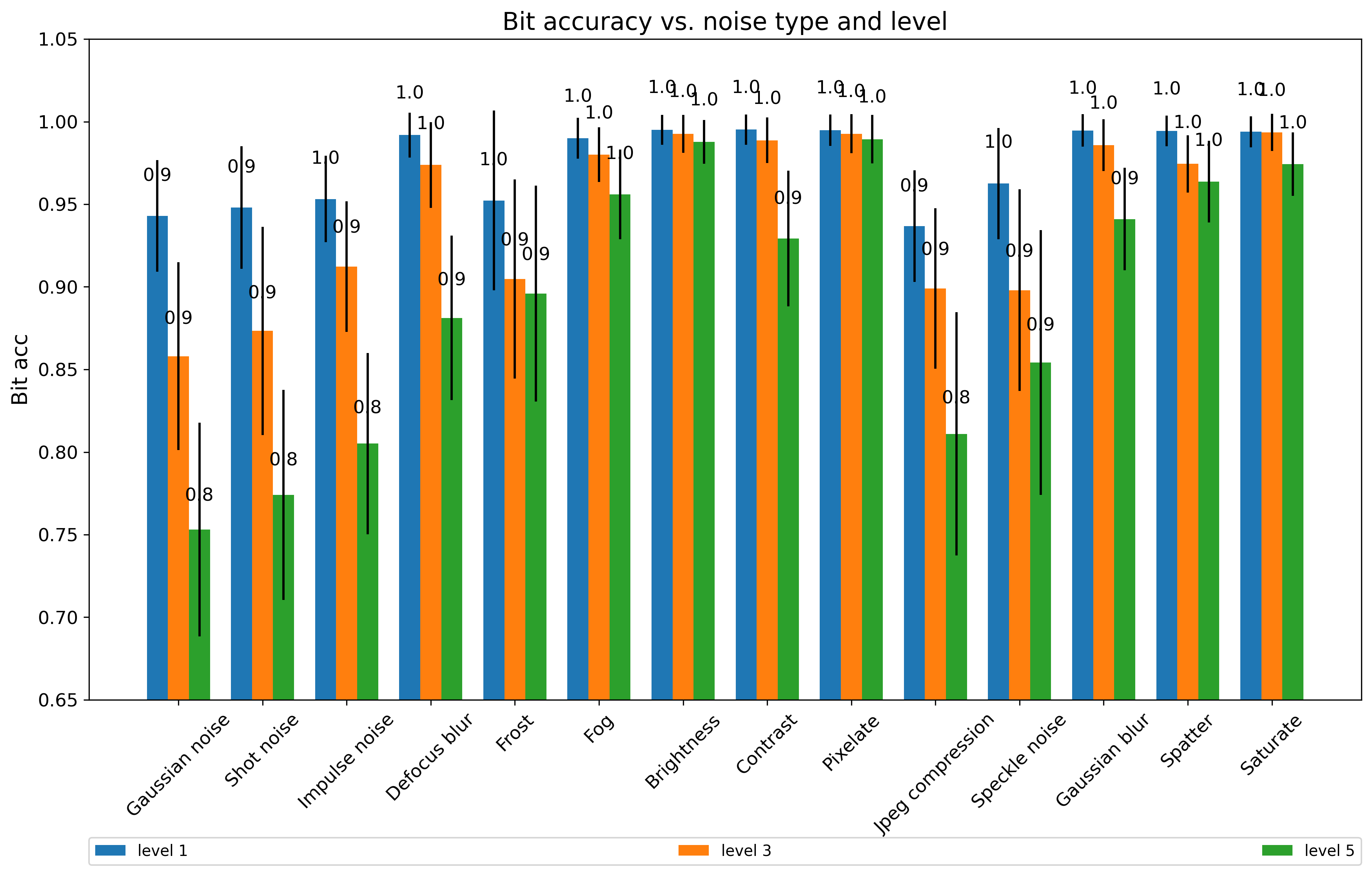}
    \caption{RoSteALS secret recovery performance breakdown for noise types and severity levels .}
    \label{fig:netc_levels}
\end{figure*}

\section{Perturbations}
\Cref{fig:netc} shows examples of 14 ImageNet-C perturbations used in our work. Note that there are 19 perturbations in ImageNet-C in total, we exclude 5 of them which are too slow to be included in training. Each perturbation has 5 levels of severity and its performance breakdown per level is shown in \Cref{fig:netc_levels}. RoSteALS is most sensitive to degradation due to Gaussian, shot, impulse and speckle noises as well as jpeg compression; while being most resilient to brightness, pixelate and saturation effects.

\section{More qualitative results}
\begin{figure*}[t!]
    \centering
    \includegraphics[width=1.0\linewidth,trim=0cm 0cm 0cm 0cm,clip,height=24cm]{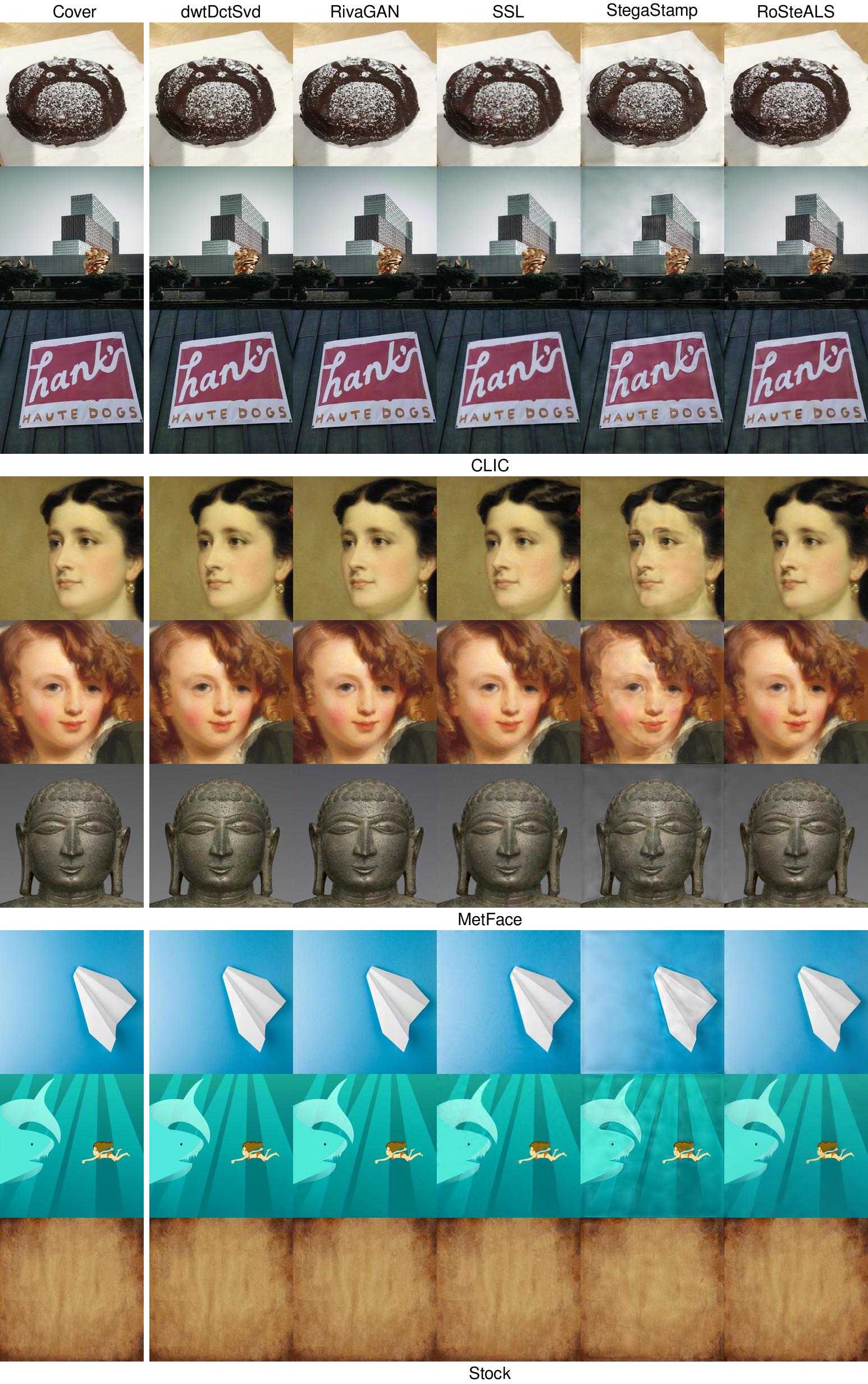}
    \vspace{-2em}
    \caption{Stego images generated from several covers and a fixed secret. RoSteALS has better image quality than StegaStamp and perceptually comparable with other methods.}
    \label{fig:qualitative}
\end{figure*}
\Cref{fig:qualitative} shows more qualitative examples of stego images created by RoSteALS and other baselines. The artifacts on StegaStamp generated images are perceptually visible, as if the image is covered with a transparent layer of fog. RoSteALS performance is comparable with other methods.

\begin{figure*}[t!]
    \centering
    \includegraphics[width=1.0\linewidth,trim=0cm 0cm 0cm 0cm,clip,height=24cm]{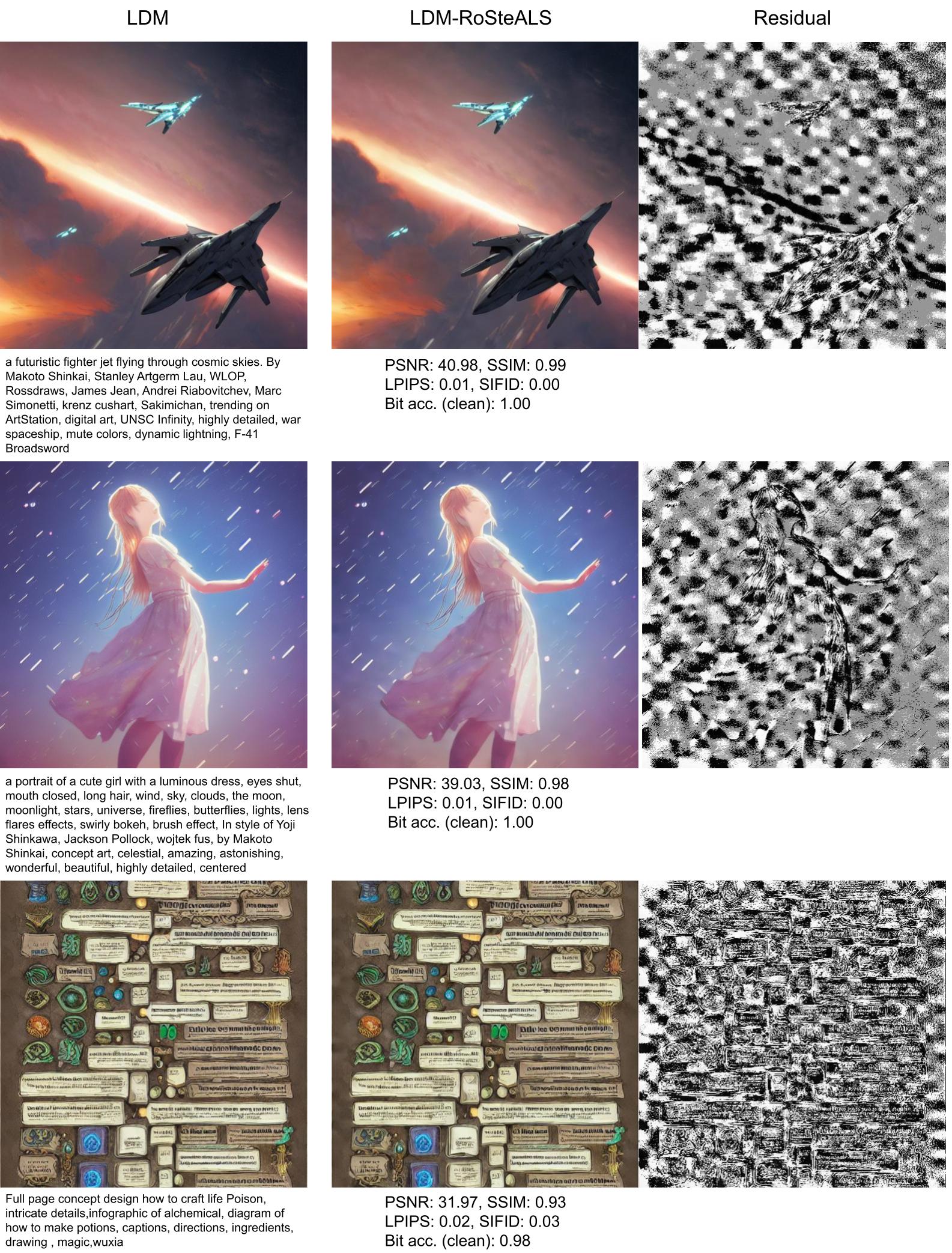}
    \vspace{-2em}
    \caption{Text-based steganography with LDM-RoSteALS. (Left)- 512x512 images sampled from Stable Diffusion using the given prompts. (Middle) - Stegos with secret word ``RoSteALS" injected. (Right) - Residual image scaled to [0,255] range. }
    \label{fig:sdeg}
\end{figure*}

\Cref{fig:sdeg} depicts several examples of our novel text-based steganography application. We note the glimpse of semantic objects visible in the residual image, however these artifacts are inevitable around the strongest edges during image generation and do not represent the secret artifacts to be picked up by the secret decoder (c.f. \Cref{fig:gradcam} in the main paper). 
\end{appendices}
\end{document}